\documentclass{article}
\usepackage{ijcai26}

\usepackage{times}
\usepackage{soul}
\usepackage{url}
\usepackage[hidelinks]{hyperref}
\usepackage[utf8]{inputenc}
\usepackage[small]{caption}
\usepackage{natbib}
\usepackage{graphicx}
\usepackage{amsmath}
\usepackage{amsthm}
\usepackage{xcolor}
\usepackage{booktabs}
\usepackage{algorithm}
\usepackage{algorithmic}
\usepackage[switch]{lineno}

\usepackage{amssymb}
\usepackage{bm}                        
\usepackage{array}
\usepackage{multirow}
\usepackage{tabularx}
\usepackage{subcaption}

\usepackage{booktabs}

\usepackage{ragged2e} 

\usepackage{graphicx}
\usepackage{xcolor}
\usepackage[most]{tcolorbox}

\usepackage{booktabs}
\usepackage{multirow}
\usepackage{colortbl}
\usepackage{xcolor}
\usepackage{booktabs}
\usepackage{makecell} 

\definecolor{tpgreen}{RGB}{221,240,203} 

\tcbset{
  evalbox/.style={
    width=\textwidth,
    colback=white,
    colframe=black!70,
    boxrule=0.8pt,
    arc=2mm,
    left=6pt,right=6pt,top=6pt,bottom=6pt,
  }
}

\title{HIME: Mitigating Object Hallucinations in LVLMs via Hallucination Insensitivity Model Editing}

\author{
Ahmed Akl$^{1,2}$
\and
Abdelwahed Khamis$^2$\and
Ali Cheraghian$^{3}$\and \\
Zhe Wang$^1$\and
Sara Khalifa$^3$\And
Kewen Wang$^1$\\
\affiliations
$^1$School of Information and Communication Technology, Griffith University, Australia\\
$^2$Data61, CSIRO, Australia\\
$^3$School of Engineering, Macquarie University, Sydney, Australia\\
$^4$School of Information Systems, Queensland University of Technology, Australia\\
\emails
ahmed.akl@griffithuni.edu.au,
abdelwahed.khamis@data61.csiro.au,
ali.cheraghian@mq.edu.au,
zhe.wang@griffith.edu.au,
sara.khalifa@qut.edu.au,
k.wang@griffith.edu.au
}

\begin{document}
\maketitle

\begin{abstract}

Large Vision–Language Models (LVLMs) have demonstrated impressive multimodal understanding capabilities, yet they remain prone to object hallucination, where models describe non-existent objects or attribute incorrect factual information, raising serious concerns for reliable real-world deployment. While fine-tuning is a commonly adopted mitigation strategy, its high computational cost and practical difficulty motivate the need for training-free alternatives, among which model editing has recently emerged as a promising direction. However, indiscriminate editing risks disrupting the rich implicit knowledge encoded in pre-trained LVLMs, leading to a fundamental question: how much intervention is necessary at each layer to suppress hallucinations while preserving pre-trained knowledge?

To address this question, we present a systematic analysis of LVLM decoders built on three widely used large language model backbones—Qwen, LLaMA, and Vicuna—revealing clear layer-wise differences in susceptibility to object hallucination. Building on these insights, we introduce the Hallucination Insensitivity Score (HIS), a principled metric that quantifies each layer’s sensitivity to hallucination and provides guidance for targeted intervention. Leveraging HIS, we propose Hallucination Insensitivity Model Editing (HIME), a simple yet effective layer-adaptive weight editing approach that selectively modifies latent features to suppress hallucinations while preserving pre-trained knowledge. Extensive experiments demonstrate that HIME reduces hallucinations by an average of 61.8\% across open-ended generation benchmarks, including CHAIR, MME, and GPT-4V-aided evaluation, without introducing additional parameters, inference-time latency, or computational overhead.

\end{abstract}

\begin{figure}[t!]
    \centering
\includegraphics[width=1\linewidth]{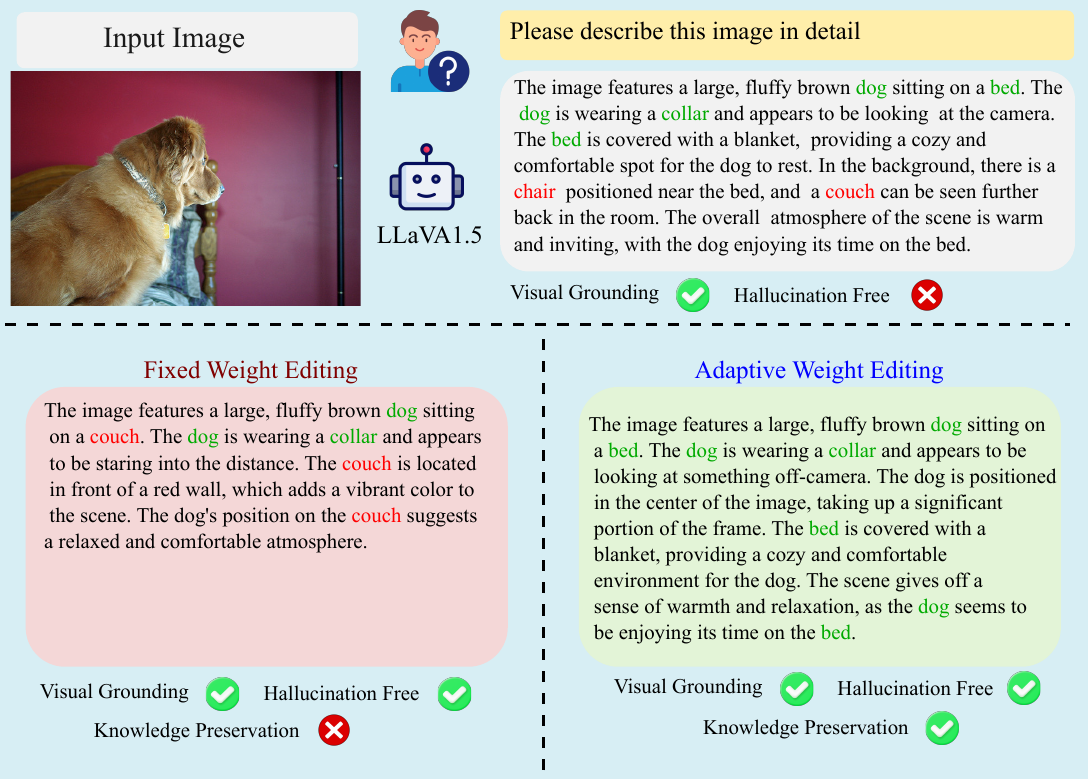}
    \vspace{-1em}
    \caption{Illustrates the phenomenon of knowledge distortion because of the fixed weight editing. On top, LVLM e.g. LLaVA-1.5 hallucinates non-existent objects, \textcolor{red}{chair, couch}, frequently co-occurring to bed. Left bottom, Fixed Weight Editing reduce hallucinated object, yet drop existent objects, \textcolor{green}{bed}. Right bottom, Our approach, HIME, mitigate hallucination, \textcolor{red}{chair, couch}, and preserve the pre-trained knowledge, \textcolor{green}{bed}.}
    \label{fig:hime_editing}
\end{figure}

\section{Introduction}
Object hallucination remains a fundamental obstacle to the reliable deployment of Large Vision--Language Models
(LVLMs) in real-world settings~\cite{LiuYeZou2024reducing}. While hallucination in Large Language Models (LLMs)
typically refers to the generation of unsupported or incorrect statements, hallucination in LVLMs is often
triggered by \emph{cross-modal misalignment}---the model produces objects or attributes that are not grounded
in the visual input~\cite{Ji2023survey,Zhang2023siren,Shi2023trusting}. Such failures can compromise the
trustworthiness of multimodal systems in safety-critical or high-stakes scenarios, motivating the development
of effective mitigation strategies~\cite{Wang2023mitigating,Zhao2023beyond,huang2024opera,Yue2024less}.

Existing mitigation methods generally fall into two categories: \textbf{(i)} fine-tuning-based approaches
(e.g., preference optimisation and supervised adaptation)~\cite{xiao2025detecting,yu2024rlhf}, and
\textbf{(ii)} training-free strategies~\cite{leng2024mitigatingvcd,wang-etal-2024-mitigating,yang2025nullu}.
Although fine-tuning can be highly effective, it often requires curated supervision, careful engineering, and
substantial computation, which limits its practicality for rapid deployment and model updates. This has fuelled interest in training-free alternatives, including decoding-time interventions
(e.g., contrastive/auxiliary decoding) ~\cite{chen2024halc,leng2024mitigatingvcd,deng2024seeing} and offline
weight-editing methods~\cite{yang2025nullu,zhuang2025vasparse,wang2025mint}. However, decoding-time methods typically incur additional inference cost and latency due to extra generation paths. Knowledge editing in LLMs has shown strong effectiveness; however, recent studies indicate that it can introduce undesirable side effects—such as knowledge distortion and knowledge conflict \cite{li2023unveiling} —which remain under-explored in LVLM weight editing. As shown in Fig\ref{fig:hime_editing}, LlaVA-1.5 model suffer from object hallucination, incorrectly describing non-existent objects, (e.g. \textit{``couch'', and ``chair''}) that are frequently co-occurring to the object (\textit{``bed''}). While Nullu \cite{yang2025nullu} applies a hard uniform weight editing that successfully removes the hallucinated object \textit{``chair''}, it does so at the cost of suppressing the existent object \textit{``bed''}, indicating a distortion of factual visual knowledge. Therefore, uniform model editing introduces a potential risk: \emph{indiscriminate parameter modifications that can disrupt the implicit knowledge encoded in pre-trained LVLMs}. This raises a central question: \textbf{to what extent should we intervene in each layer to suppress hallucination while preserving the model's pre-trained knowledge?}

In this work, we show that object hallucination in Large Vision–Language Models (LVLMs) is not a uniform decoder-level phenomenon, but instead exhibits strong and systematic variation across layers. By conducting a layer-wise analysis of widely used LVLMs, LLaVA-1.5, QWen3-VL-8B-Instruct, QWen2-VL-8B-Instruct, MiniGPT-4, and mPLUG-Owl2—we uncover depth-dependent patterns in hallucination susceptibility that are consistent across architectures. This observation challenges the common practice \cite{yang2025nullu} of uniformly editing or regularising all decoder layers, which implicitly assumes equal contribution to hallucination and risks indiscriminately perturbing grounded visual representations and pre-trained semantic knowledge.

Building on these insights, we introduce the Hallucination Insensitivity
Score (HIS), a principled layer-level metric that quantifies how sensitive a layer is to hallucination and
therefore provides guidance for targeted intervention. Leveraging HIS, we propose
Hallucination Insensitivity Model Editing (HIME), a simple yet effective layer-adaptive weight
editing approach that selectively suppresses hallucination-related latent directions while preserving
pre-trained knowledge. Concretely, we derive attention-guided features for truthful and hallucinated samples,
estimate their contrastive discrepancy, and extract a low-rank hallucination subspace via Singular Value Decomposition (SVD). We then apply a
layer-adaptive weighted projection---controlled by Hallucination Insensitivity Score 
to edit MLP weights offline, yielding an
edited LVLM that can be directly reloaded for inference with \emph{zero} additional parameters, latency, or
computational overhead.

Extensive experiments show that HIME reduces hallucinations by an average of $61.8\%$ across open-ended
generation benchmarks, including CHAIR, MME, and GPT-4V-aided evaluation, while maintaining overall model utility.

Our main contributions are: 
\begin{enumerate}

    \item We perform a layer-wise analysis of LVLM decoders and reveal that object hallucination LVLMs exhibit pronounced, depth-dependent susceptibility patterns, with certain layers consistently more prone to hallucination across architectures.

    \item We introduce the Hallucination Insensitivity Score (HIS), a layer-level metric that quantifies hallucination susceptibility from internal decoder representations, and leverage it to develop Hallucination Insensitivity Model Editing (HIME), a training-free, layer-adaptive weight editing method for hallucination mitigation.

    \item Across three LVLM backbones (LLaVA-1.5, MiniGPT-4, and mPLUG-Owl2) and multiple benchmarks, HIME consistently outperforms prior decoding- and editing-based approaches. HIME reduces object hallucination by an average of 61.8\% on open-ended generation (CHAIR), while maintaining or improving BLEU and MME perception scores, with zero additional parameters, training, or inference-time overhead.

\end{enumerate}

\begin{figure*}[t!]
    \centering
    \includegraphics[width=1\linewidth]{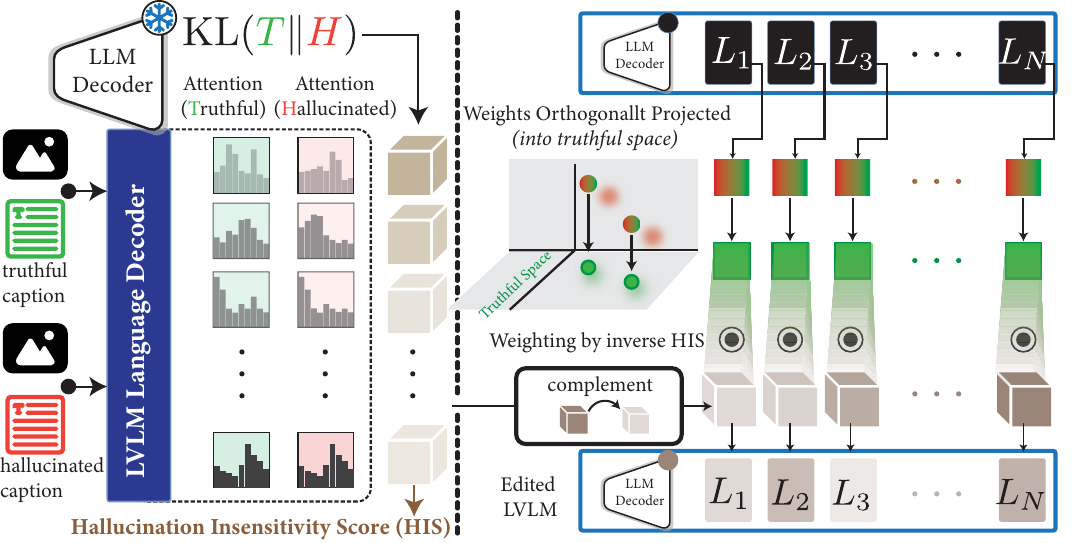}
    \caption{\textbf{HIME hallucination mitigation framework.} Given an image with truthful and hallucinated captions, the Hallucination Insensitivity Score (HIS)  is derived by contrasting attention distributions via KL divergence. During model editing , model weights are orthogonally projected into the truthful subspace and weighted by inverse HIS.
     The resulting Edited LVLM Decoder emphasises truthful representations while reducing hallucination sensitivity across layers.}
    \label{fig:game_main}
\end{figure*}

\section{Related Work}

\noindent\textbf{Hallucination Mitigation in LVLMs:}
The recent adoption of advanced open-source large language models, including Llama \cite{touvron2023llama}, Vicuna \cite{zheng2023lmsys}, and Qwen \cite{bai2023qwen} has markedly enhanced LVLM performance, allowing them to tackle increasingly complex vision–language tasks. However, these models suffer from object hallucination when generate content inconsistent with the image (i.e., ungrounded), compounded by next-token training objectives that do not penalise unfaithful text. Benchmarks such as CHAIR \cite{rohrbach2018object} has highlighted persistent grounding failures. 

\noindent Various mitigation strategies have been proposed to mitigate hallucination in vision-language models. Mitigation strategies can be classified into three categories. (1) Contrastive Decoding, originally introduced in NLP \cite{li2022contrastive}, has demonstrated effectiveness in reducing hallucination in LVLMs. The adoption of this technique has motivated a range of contrastive strategies such as VCD~\cite{leng2024mitigatingvcd}, and DoLa~\cite{chuang2024dola}. While they achieve good performance, they incur heavy computational costs, particularly during training and fine-tuning.
 
\noindent In contrast to our training-free approach. (2) Feature Steering: contrast between the internal embeddings of positive and negative samples, and then edit the feature during the test-time to adjust the model's response e.g. VTI~\cite{liu2024reducing}.

Although training-free, it lacks input adaptivity, as fixed shifts may be suboptimal across different scenes or prompts. (3) Null-Space Projection: leverages the null space projection for model editing. Nullu~\cite{yang2025nullu} mitigates hallucinations by projecting contrastive latent features into the null space (HalluSpace), then orthogonalising the model weights to the null space. While effective, it applies fixed-weight editing to all layers, distorting the implicit model information. In contrast, our approach account for layer-wise susceptibility and applies layer-adaptive interventions that remove hallucination and preserve the pre-trained knowledge.

\section{Method}
\label{sec:method}

The overall framework is illustrated in Figure~\ref{fig:game_main}. We first introduce preliminary concepts and notation used throughout the paper. We then present the Hallucination Insensitivity Score (HIS), which analyzes attention sensitivity across layers in recent LVLMs. Finally, we introduce Hallucination Insensitivity Model Editing (HIME), a method designed to mitigate object hallucination in LVLMs.

\subsection{Preliminary}

\noindent\textbf{Vision-Language Alignment.} The input to a vision-language foundation model (LVLM) consists of an image
$\mathbf{I}^{(i)} \in \mathbb{R}^{H \times W \times C}$ and a textual query
$s^{(i)}$. A vision encoder (e.g., ViT~\cite{dosovitskiy2021an}, CLIP~\cite{pmlr-v139-radford21a})
extracts visual features from $\mathbf{I}^{(i)}$, which are then mapped into the
language model embedding space via a vision--language alignment module
(e.g., Q-Former~\cite{10.5555/3618408.3619222} or a linear projection). This
produces a sequence of $N$ visual tokens
\[
\mathbf{X}^{(i)} =
[\mathbf{x}^{(i)}_0, \mathbf{x}^{(i)}_1, \dots, \mathbf{x}^{(i)}_{N-1}],
\quad \mathbf{x}^{(i)}_n \in \mathbb{R}^d .
\]
The textual query $q^{(i)}$ is tokenized into a sequence of $M$ text tokens
\[
\mathbf{T}^{(i)} =
[\mathbf{t}^{(i)}_0, \mathbf{t}^{(i)}_1, \dots, \mathbf{t}^{(i)}_{M-1}],
\quad \mathbf{t}^{(i)}_m \in \mathbb{R}^d .
\]
The combined multimodal input is formed by concatenation,
$[\mathbf{X}^{(i)}, \mathbf{T}^{(i)}] \in \mathbb{R}^{J \times d}$,
where $J = N + M$.

\vspace{0.2cm}
\noindent\textbf{Model Forwarding.}
The concatenated token sequence is processed by the transformer backbone of
the LVLM. Let $L$ denote the total number of transformer layers, and let
$\mathbf{z}_{\ell,j}^{(i)} \in \mathbb{R}^d$ denote the hidden representation of
token index $j \in \{0,\dots,J-1\}$ at layer $\ell \in \{1,\dots,L\}$ for sample
$i$. The model produces a set of contextualized token representations
\begin{equation}
\left\{ \mathbf{z}_{\ell,j}^{(i)} \right\}_{\ell=1,j=0}^{L,\,J-1}
= f_{\theta}^{\text{LVLM}}\!\left(\mathbf{I}^{(i)}, s^{(i)}\right).
\end{equation}
These hidden states jointly encode visual and linguistic context and form the
basis for downstream reasoning and response generation.

\vspace{0.2cm}
\noindent\textbf{Response Generation.}
Following the forward pass, response generation is performed in an
autoregressive manner. At decoding step $t$, the probability distribution over
the next token $y_{t+1}^{(i)}$ is computed from the final-layer hidden
representation corresponding to the current decoding position,
\begin{equation}
P\!\left(y_{t+1}^{(i)} \mid y_{1:t}^{(i)}, \mathbf{z}_{L,0:J-1}^{(i)}\right)
= \text{softmax}\!\left( \mathbf{W}_o \, \mathbf{z}_{L,j_t}^{(i)} \right),
\end{equation}
where $j_t$ indexes the token position at step $t$, and
$\mathbf{W}_o \in \mathbb{R}^{V \times d}$ is the output projection matrix for a
vocabulary of size $V$. Decoding proceeds until an end-of-sequence token is
generated or a predefined maximum length is reached.

\begin{figure}[t!]
    \centering

\includegraphics[width=1\linewidth]{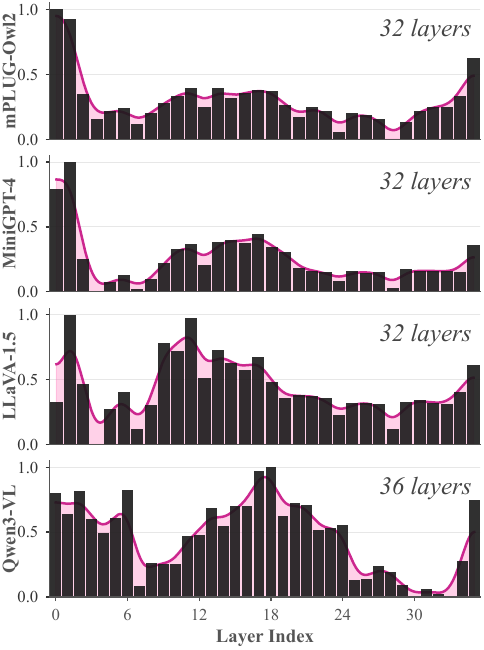}

    \caption{Layer-wise distribution of the Hallucination Insensitivity Score (HIS) across four VLM backbones. The y-axis reports the HIS value (higher indicates greater robustness to hallucinations). Across four VLM decoders, we observe a highly structured, non-uniform depth profile. Across architectures, mid-depth layers exhibit consistently higher robustness, while late/early decoder layers show pronounced sensitivity to hallucinations. This recurring pattern suggests targeted editing of specific depth regions rather than global model intervention.}
    \label{fig:HIS_family}
\end{figure}

\subsection{LVLM Attention Behaviour}

\noindent Following \cite{yang2025nullu} strategy, we utilise the contrastive dataset LURE \cite{zhou2024analyzing} that contain vision-language pairs in which each image is associated with two different captions: one hallucinated and one truthful ground truth. We denote \(x_{i}^{+}\) as the ground-truth sample that precisely describes the image contents, and \(x_{i}^{-}\) as the hallucinated sample whose caption contains hallucinated descriptions. The whole dataset can be denoted as \(\mathcal{D} = (x_{i}^{+}, x_{i}^{-})_{i=1}^{N}\).

\noindent GPT-3.5 is employed to transform the accurate captions into the hallucinated counterparts, as described in \cite{zhou2024analyzing}. These transformations are driven by factors implicated in object hallucination co-occurrence statistics, uncertainty of object existence, and positional placement within the caption. Thereafter, we obtain paired samples by replacing objects in the ground-truth description with their most probable hallucinated counterparts, resulting in an image paired with both the ground-truth and the hallucinated descriptions.

\noindent\textbf{LVLM Hallucination Insensitivity Score (HIS).} LVLMs rely on a Transformer-based LLM decoder that generates text using attention mechanisms. Consequently, the importance of each token depends on how much attention it receives across the layers during decoding. To systematically investigate how LVLMs respond to the input tokens during response generation, we introduce a \textit{Hallucination Insensitivity Score} which is a distributional distance measure that quantifies and tracks, across layer, the degree of attention sensitivity between truthful and hallucinated tokens during the decoding process. This layer-adaptive metric of hallucination susceptibility is used to guide targeted, selective interventions.

Formally, we fed both the truthful, \(x_{i}^{+}\),  and hallucinated, \(x_{i}^{-}\), samples to the LVLM, and calculate the attention matrix over each head for both of them using the following equation:

\begin{equation}\label{eq:head_attn}
\mathbf{A}^{(h)} \;=\; 
\mathrm{Softmax}\!\left(
  \frac{\mathbf{Q}^{(h)} {\mathbf{K}^{(h)}}^{\!\top}}{\sqrt{d_k/H}}
\right)\mathbf{V}^{(h)} 
\end{equation}

\noindent where \(h\) is the head number,  \(d_k\) represents embedding dimension, \(H\) is the numbrer of attention head, and  \(\mathbf{Q}, \mathbf{K}, \mathbf{V}\) represent Query, Key, and Value, respectively.

\noindent For simplicity, we average attention scores across heads using $\bar{\mathbf{A}}^{+}_{l}=\frac{1}{H}\sum_{h=1}^{H}\mathbf{A}^{+}_{l,h}, \;\; \bar{\mathbf{A}}^{+}_{l}\in\mathbb{R}^{J\times J}$, and $J$ is length of all tokens, to derive a single layer-level attention matrix over all tokens, $\bar{\mathbf{A}}^{+}_{l}$ and similarly, for $\bar{\mathbf{A}}^{-}_{l}$, where $\ell \in \{1,\dots,L\}$. Because each token distributes attention across the entire sequence, we require a representation that preserves meaningful token-level variability. One option is to average attention over tokens; however, this implicitly weights all tokens equally, despite substantial disparities in their attention magnitudes. To address this, we propose a distributional divergence that characterizes the overall attention behaviour. Specifically, for each layer $l$, we flatten 
$\bar{\mathbf{A}}^{+}_{l}$ and $\bar{\mathbf{A}}^{-}_{l}$ to get a vector of attention scores $\mathbf{a}^{+}_{l}=\mathrm{vec}\!\left(\bar{\mathbf{A}}^{+}_{l}\right)\in\mathbb{R}^{J^{2}}$, and $\mathbf{a}^{-}_{l}$, then map both of them into a histogram  $p^+_{\ell} = [a^+_{0,l}, a^+_{1,l},\cdots, a^+_{B,l}]$ and $q^-_{\ell} = [a^-_{0,l}, a^-_{1,l},\cdots, a^-_{B,l}]$, where $p_i, q_i \in \mathbb{R}^B$ with $B$ bins. 

We then quantify layer-wise hallucination sensitivity via KL divergence, where $b$ indexes histogram bins.

\begin{equation}\label{eq:kl_hist_layer}
\mathrm{HIS}_{\ell}
\;=\;
D_{\mathrm{KL}}(p_{}\ell\,\|\,q_{\ell})
\;=\;
\sum_{b=1}^{B} p_{l,b}\log\frac{p_{l,b}}{q_{l,b}}.
\end{equation}

We introduce \textit{Hallucination Insensitivity Score} (HIS) to quantify, at each decoder layer, how separable the attention behaviour is between truthful and hallucinated generations. A larger HIS implies a greater divergence between the two attention distributions, indicating that the layer can more clearly discriminate hallucinated from truthful outputs. Conversely, a smaller HIS reflects highly similar distributions, suggesting limited sensitivity (i.e., potential confusion) and therefore highlighting the layer as a suitable target for focused intervention or repair, see Figure \ref{fig:HIS_family}.

\subsection{Hallucination Insensitivity Model Editing}

\noindent The internal outputs such as hidden states and self-attention maps are contextualised representations that jointly capture the model's reasoning and generation ability, but they also entangle signals related to language hallucination \cite{vaswani2017attention}. Consequently, relying on hidden states alone is often insufficient for isolating the hallucination subspaces \cite{yang2025nullu}. We therefore leverage layer-wise attention to derive more informative and guided hidden representations, which we then use to identify and suppress hallucination subspaces.

\noindent\textbf{Layer-wise attention-guided representations.} Formally, for each layer $\ell$ in the LVLM backbone, where $\ell \in \{1,\dots,L\}$, we extract contextual embedding features for both truthful and hallucinated samples, denoted as $E^{+}_{\ell}$ and $E^{-}_{\ell} \in\mathbb{R}^{N \times J\times D}$, where $N$ number of samples, $J$ sequence length, and $D$ MLP embedding dimension. Unlike prior methods~\cite{yang2025nullu} that rely solely on hidden states, we additionally leverage layer-wise attention to obtain more informative and guided representations. Specifically, for each contrastive sample, we compute the per-layer attention matrices $A^{+}_{\ell}$ and $A^{-}_{\ell}$ using Eq.~\ref{eq:head_attn}. We then normalise the attention per head and average across heads to obtain a single layer-level attention matrix $\bar{A}^{+}_{\ell}$ and $\bar{A}^{-}_{\ell} \in\mathbb{R}^{J \times J}$ for both truthful and hallucinated samples, respectively.

To derive a one-dimensional positional importance distribution, we aggregate $\bar{A}_{\ell}$ over the \emph{key} dimension: $\boldsymbol{\pi}_{\ell}[q] \;=\; \frac{1}{J}\sum_{k=1}^{J}\bar{A}_{\ell}[q,k]$ Consequently, $\boldsymbol{\pi}_{\ell}\in\mathbb{R}^{J}$ is a 1D positional distribution of length $J$ that assigns,
for each query position, its average attention mass allocated across all keys at layer $\ell$.

\noindent Using the hidden embeddings at layer $\ell$, $E^{+}_{\ell}$ and $E^{-}_{\ell}$, we combine them with the
positional attention distribution $\mathbf{p}_{\ell}$ to obtain an attention-weighted expectation of the
hidden states. Specifically, we compute
$\mathbf{Z}^{+}_{\ell} \;=\; {\mathbf{p}^{+}_{\ell}} E^{+}_{\ell}$ and
$\mathbf{Z}^{-}_{\ell} \;=\; {\boldsymbol{\pi}^{-}_{\ell}} E^{-}_{\ell}$,
yielding attention-guided features for truthful and hallucinated generations, respectively, where
$\mathbf{Z}^{+}_{\ell},\mathbf{Z}^{-}_{\ell}\in\mathbb{R}^{N \times D}$, where $D$ denotes the LLM embedding dimension and $N$ is the number of samples.

To characterise the latent discrepancy between truthful and hallucinated embeddings in the feature space, we
compute a layer-wise difference matrix at layer $\ell$:
\begin{equation}\label{eq:diff_matrix}
\mathbf{Z}_{\ell} \;=\; \mathbf{Z}^{+}_{\ell} - \mathbf{Z}^{-}_{\ell},
\qquad \mathbf{Z}_{\ell} \in \mathbb{R}^{N \times D}.
\end{equation}
\noindent
We then apply singular value decomposition (SVD) to obtain a low-rank factorisation of $\mathbf{Z}_{\ell}$,
thereby capturing the dominant directions that separate truthful from hallucinated features. These principal
directions define a compact subspace that can be leveraged to suppress hallucination-related components in
the model representations:
\begin{equation}\label{eq:svd-El}
\mathbf{Z}_{\ell}
= \mathbf{U}_{\ell}\,\boldsymbol{\Sigma}_{\ell}\,\mathbf{V}_{\ell}^{\top},
\qquad
\mathbf{U}_{\ell}\in\mathbb{R}^{N\times N},\;
\mathbf{V}_{\ell}\in\mathbb{R}^{D\times D}.
\end{equation}
\noindent
Here, $\boldsymbol{\Sigma}_{\ell}$ is a diagonal matrix whose entries are the singular values sorted in
descending order. We select the top-$k$ right singular vectors
$\mathbf{v}^{\ell}_{1},\dots,\mathbf{v}^{\ell}_{k}$ (the first $k$ columns of $\mathbf{V}_{\ell}$), which
capture the most salient directions of discrepancy between truthful and hallucinated features. We stack these
vectors to form $\mathbf{V}_{\ell,k}\in\mathbb{R}^{D\times k}$ and regard their span as the \emph{hallucination
subspace} at layer $\ell$.

\noindent\textbf{Weighted model editing.}
Let $\mathbf{V}_{\ell,k}\in\mathbb{R}^{D\times k}$ denote the top-$k$ right singular vectors at layer $\ell$,
and define the hallucination projector
$\mathbf{V}_{\ell,k}\mathbf{V}_{\ell,k}^{\top}$.
We modulate the intervention strength using the complement score $HIS^{c}_{\ell}\in[0,1]$, yielding the
\emph{weighted null-space operator}
\begin{equation}\label{eq:weighted_null}
\mathbf{N}_{\ell} \;=\; \mathbf{I}-\text{HIS}^{c}_{\ell}\mathbf{V}_{\ell,k}\mathbf{V}_{\ell,k}^{\top}.
\end{equation}
\noindent
Compared to prior work~\cite{yang2025nullu}, which applies the full projection
$\mathbf{I}-\mathbf{P}_{\ell}$ and may induce abrupt parameter changes, Eq.~\ref{eq:weighted_null} provides a
smooth interpolation from \emph{no editing} ($HIS^{c}_{\ell}=0$) to \emph{full projection} ($HIS^{c}_{\ell}=1$)

We then edit the MLP weights by applying $\mathbf{N}_{\ell}$ on the appropriate side:
\begin{equation}\label{eq:weighted_edit_mlp}
\mathbf{W}^{\mathrm{ed}}_{\ell} \;=\; \mathbf{N}_{\ell}\mathbf{W}^{\mathrm{org}}_{\ell},
\qquad
\end{equation}
\noindent
At inference time, we replace $\mathbf{W}^{\mathrm{org}}_{\ell}$ with $\mathbf{W}^{\mathrm{ed}}_{\ell}$, enabling
hallucination mitigation without introducing additional parameters, latency, or computational overhead, since
the edited weights can be directly reloaded into the LVLM.

\begin{table*}[t]
\centering

\setlength{\tabcolsep}{4pt}
\renewcommand{\arraystretch}{1.10}
\scalebox{0.75}{
\begin{tabular}{l !{\vrule width 0.6pt}ccc !{\vrule width 0.6pt}ccc !{\vrule width 0.6pt}ccc}
\toprule
\multirow{2}{*}{\textbf{Method}} &
\multicolumn{3}{c}{\textbf{LLaVA-1.5}} &
\multicolumn{3}{c}{\textbf{MiniGPT-4}} &
\multicolumn{3}{c}{\textbf{mPLUG-Owl2}} \\
\cmidrule(lr){2-4}\cmidrule(lr){5-7}\cmidrule(lr){8-10}
& $\text{CHAIR}_s \downarrow$ & $\text{CHAIR}_i \downarrow$ & BLEU $\uparrow$
& $\text{CHAIR}_s \downarrow$ & $\text{CHAIR}_i \downarrow$ & BLEU $\uparrow$
& $\text{CHAIR}_s \downarrow$ & $\text{CHAIR}_i \downarrow$ & BLEU $\uparrow$ \\
\midrule
Greedy
& $20.40_{\pm 2.80}$ & $7.08_{\pm 0.33}$ & $15.72_{\pm 0.10}$
& $32.40_{\pm 2.20}$ & $12.20_{\pm 0.42}$ & $14.57_{\pm 0.11}$
& $22.90_{\pm 0.90}$ & $8.62_{\pm 0.11}$ & $15.01_{\pm 0.24}$ \\
Beam Search \cite{freitag2017beam}
& $19.50_{\pm 2.30}$ & $6.84_{\pm 0.79}$ & $15.99_{\pm 0.14}$
& $30.10_{\pm 0.30}$ & $11.87_{\pm 0.37}$ & $15.35_{\pm 0.24}$
& $20.30_{\pm 0.70}$ & $6.92_{\pm 0.19}$ & $15.43_{\pm 0.53}$ \\
DoLa \cite{chuang2024dola}
& $20.20_{\pm 2.80}$ & $6.55_{\pm 0.70}$ & $15.58_{\pm 0.62}$
& $31.00_{\pm 2.10}$ & $12.15_{\pm 0.89}$ & $14.54_{\pm 0.12}$
& $22.40_{\pm 0.80}$ & $7.36_{\pm 0.34}$ & $15.13_{\pm 0.26}$ \\
OPERA \cite{huang2024opera}
& $17.50_{\pm 0.50}$ & $6.07_{\pm 0.33}$ & $16.02_{\pm 0.12}$
& $29.70_{\pm 0.40}$ & $11.90_{\pm 0.36}$ & $14.82_{\pm 0.05}$
& $20.67_{\pm 0.72}$ & $6.40_{\pm 0.18}$ & $15.41_{\pm 0.18}$ \\
VCD \cite{leng2024mitigatingvcd}
& $20.30_{\pm 1.10}$ & $7.28_{\pm 1.40}$ & $14.53_{\pm 0.01}$
& $29.20_{\pm 0.80}$ & $11.69_{\pm 0.11}$ & $14.42_{\pm 0.01}$
& $22.80_{\pm 0.80}$ & $8.67_{\pm 0.17}$ & $15.14_{\pm 0.13}$ \\
Woodpecker \cite{yin2024woodpecker}
& $23.85_{\pm 4.62}$ & $7.50_{\pm 0.90}$ & $15.01_{\pm 0.05}$
& $30.40_{\pm 1.80}$ & $12.00_{\pm 0.85}$ & $15.00_{\pm 0.07}$
& $26.33_{\pm 1.98}$ & $9.43_{\pm 0.83}$ & $15.03_{\pm 0.20}$ \\
LURE \cite{zhou2024analyzing}
& $19.48_{\pm 2.35}$ & $6.50_{\pm 0.33}$ & $15.97_{\pm 0.01}$
& $27.88_{\pm 2.25}$ & $10.20_{\pm 0.85}$ & $15.03_{\pm 0.11}$
& $21.27_{\pm 0.06}$ & $7.67_{\pm 0.16}$ & $15.65_{\pm 0.15}$ \\
HALC \cite{chen2024halc}
& $16.90_{\pm 2.56}$ & $5.72_{\pm 0.56}$ & $16.02_{\pm 0.04}$
& $25.20_{\pm 2.20}$ & $9.42_{\pm 0.41}$ & $14.91_{\pm 0.31}$
& $18.80_{\pm 1.80}$ & $7.00_{\pm 0.01}$ & $15.33_{\pm 0.24}$ \\
Nullu \cite{yang2025nullu}
& $15.20_{\pm 0.60}$ & $5.30_{\pm 0.03}$ & $15.69_{\pm 0.04}$
& $21.40_{\pm 1.00}$ & $8.99_{\pm 0.36}$ & $14.81_{\pm 0.06}$
& $15.60_{\pm 1.20}$ & $5.77_{\pm 0.01}$ & $15.45_{\pm 0.01}$ \\

\textbf{HIME (our)}
& $\mathbf{13.80_{\pm 0.02}}$ & $\mathbf{4.56_{\pm 0.01}}$ & $15.82_{\pm 0.02}$
& $\mathbf{16.80_{\pm 0.01}}$ & $\mathbf{7.43_{\pm 0.02}}$ & $14.62_{\pm 0.01}$
& $\mathbf{15.40_{\pm 0.04}}$ & $\mathbf{5.19_{\pm 0.01}}$ & $\mathbf{15.81_{\pm 0.08}}$ \\
\bottomrule
\end{tabular}}
\caption{CHAIR evaluation results on MSCOCO dataset of LVLMs (LLaVA-1.5, MiniGPT-4 and mPLUG-Owl2) with different methods for mitigating OH. Lower $\text{CHAIR}_s$ and $\text{CHAIR}_i$ indicate less Object Hallucination. Higher BLEU generally represents higher captioning quality. We use 64 as the max token number in this experiment. \textbf{Bold} indicates the best result of all methods.}
\label{tab:chair_results}
\end{table*}

\begin{table}[t]
\centering
\footnotesize
\setlength{\tabcolsep}{4pt}
\renewcommand{\arraystretch}{1.05}

\resizebox{\columnwidth}{!}{%
\begin{tabular}{lccc}
\toprule
\textbf{LVLMs} & \textbf{CHAIR}$_S$$\downarrow$ & \textbf{CHAIR}$_I$ $\downarrow$ & \textbf{BLEU}$\uparrow$ \\
\midrule

\makecell[l]{QWen2-VL-8B-Instruct(QWen2)} & 20.8 & 5.36 & 11.16 \\
\textbf{HIME} & \textbf{17.2} & \textbf{4.43} & \textbf{11.30} \\
\midrule 

\makecell[l]{QWen3-VL-8B-Instruct(QWen3)} & 8.40 & 4.62 & \textbf{9.81} \\
\textbf{HIME} & \textbf{6.00} & \textbf{3.44} & 8.89 \\
\bottomrule
\end{tabular}%
}

\caption{Evaluating HIME with advanced and recent LVLMs using backbones other than Llama. Lower CHAIR$_s$, and CHAIR$_i$ is better.}
\label{tab:qwen_chair_results}
\end{table}

\begin{algorithm}[t]
\caption{HIME (Hallucination Insensitivity Model Editing)}
\label{alg:hime}
\begin{algorithmic}[1]
\STATE \textbf{Input:} Contrastive samples $\mathcal{D}=\{(x_i^{+},x_i^{-})\}_{i=1}^{N}$, LVLM $\mathcal{M}$, target layers $\mathcal{L}$, rank $k$, layer-adaptive coefficients $\{HIS^c_\ell\}_{\ell\in\mathcal{L}}$.
\STATE \textbf{Output:} Edited LVLM $\mathcal{M}^{\mathrm{ed}}$.
\FOR{$\ell \in \mathcal{L}$}
    \STATE Extract hidden embeddings $E^{+}_{\ell}, E^{-}_{\ell}$  for all $(X^{+},X^{-})\in\mathcal{D}$ using $\mathcal{M}$
    \STATE Compute layer attention $A^{+}_{\ell}, A^{-}_{\ell}\in\mathbb{R}^{J\times J}$ via Eq.~\ref{eq:head_attn}; normalise and average across heads to obtain $\bar{A}^{+}_{\ell}, \bar{A}^{-}_{\ell}$
    \STATE Compute positional attention distributions over \emph{keys} (column mean):
    $\boldsymbol{\pi}^{+}_{\ell}[k] \gets \frac{1}{J}\sum_{q=1}^{J}\bar{A}^{+}_{\ell}[q,k]$ \;and\;
    $\boldsymbol{\pi}^{-}_{\ell}[k] \gets \frac{1}{J}\sum_{q=1}^{J}\bar{A}^{-}_{\ell}[q,k]$
    \STATE Attention-weighted features:
    $\mathbf{Z}^{+}_{\ell} \gets \boldsymbol{\pi}^{+}_{\ell} E^{+}_{\ell}$,\;
    $\mathbf{Z}^{-}_{\ell} \gets \boldsymbol{\pi}^{-}_{\ell} E^{-}_{\ell}$, 
    where $\mathbf{Z}^{+}_{\ell},\mathbf{Z}^{-}_{\ell}\in\mathbb{R}^{N\times D}$
    \STATE Contrastive difference matrix: $\mathbf{Z}_{\ell} \gets \mathbf{Z}^{+}_{\ell} - \mathbf{Z}^{-}_{\ell}$
    \STATE $[\mathbf{U}_{\ell}, \boldsymbol{\Sigma}_{\ell}, \mathbf{V}_{\ell}] \gets \mathrm{SVD}(\mathbf{Z}_{\ell})$
    \STATE $\mathbf{V}_{\ell,k} \gets \mathbf{V}_{\ell}(:,1\!:\!k)$ \hfill $\triangleright$ Top-$k$ right singular vectors
    \STATE Hallucination projector: $\mathbf{P}_{\ell} \gets \mathbf{V}_{\ell,k}\mathbf{V}_{\ell,k}^{\top}$
    \STATE Weighted null operator: $\mathbf{N}_{\ell} \gets \mathbf{I} - \text{HIS}^c_{\ell}\mathbf{P}_{\ell}$
    \STATE Edit MLP weights (apply on the appropriate side):
    $\mathbf{W}^{\mathrm{ed}}_{\ell,\mathrm{up}} \gets \mathbf{N}_{\ell}\mathbf{W}^{\mathrm{org}}_{\ell,\mathrm{up}}$,\;
    $\mathbf{W}^{\mathrm{ed}}_{\ell,\mathrm{down}} \gets \mathbf{W}^{\mathrm{org}}_{\ell,\mathrm{down}}\mathbf{N}_{\ell}$
\ENDFOR
\STATE Load edited weights into $\mathcal{M}$ and return $\mathcal{M}^{\mathrm{ed}}$
\end{algorithmic}
\end{algorithm}

\section{Experimental Setup}

In this section, we briefly introduce the baselines and the standard benchmarks used to evaluate the object hallucination of LVLMs, implementation details in Appendix. We take the average of three experiments.

\subsection{Baselines and Datasets}

\begin{table*}[t]
\centering
\footnotesize
\setlength{\tabcolsep}{4.5pt}
\renewcommand{\arraystretch}{1.15}
\begin{tabular}{l l r r r r r}
\toprule
\textbf{Model} & \textbf{Method} &
\textbf{Existence} & \textbf{Count} & \textbf{Position} & \textbf{Color} & \textbf{Posters} \\
\midrule
LLaVA-1.5 & Original
& $181.67\pm2.36$ & $118.33\pm12.47$ & $104.44\pm10.39$ & $152.78\pm5.67$ & $117.23\pm4.79$ \\
LLaVA-1.5 & Nullu
& $190.00\pm4.08$ & $121.11\pm7.74$ & $105.56\pm8.30$
& \textbf{156.67$\pm$9.81} & $127.55\pm4.20$ \\
\textbf{LLaVA-1.5} & \textbf{HIME}
& \textbf{195.00$\pm$0.00} & \textbf{155.56$\pm$4.81} & \textbf{123.33$\pm$0.00}
& 151.67$\pm$5.77 & \textbf{130.27$\pm$0.59} \\
\midrule
\textbf{Model} & \textbf{Method} &
\textbf{Celebrity} & \textbf{Scene} & \textbf{Landmark} & \textbf{Artwork} & \textbf{OCR} \\
\midrule
\textbf{LLaVA-1.5} & Original
& $111.67\pm3.90$ & $144.83\pm1.50$ & $130.65\pm5.26$ & $108.92\pm2.99$ & $75.83\pm5.89$ \\
\textbf{LLaVA-1.5} & Nullu
& $115.59\pm6.60$ & $147.92\pm1.36$ & $131.66\pm1.09$
& $113.00\pm2.07$ & $121.67\pm8.25$ \\
\textbf{LLaVA-1.5} & \textbf{HIME}
& \textbf{135.59$\pm$0.51} & \textbf{154.50$\pm$0.43} & \textbf{161.67$\pm$0.14}
& \textbf{120.75$\pm$0.00} & \textbf{125.00$\pm$0.00} \\
\bottomrule
\end{tabular}
\caption{Results on MME perception-related tasks. Bold indicates the best score per column (mean $\pm$ sample SD).}
\label{tab:mme-perception}
\end{table*}

\begin{figure}[t]
  \centering
    \centering
    \includegraphics[width=\linewidth]{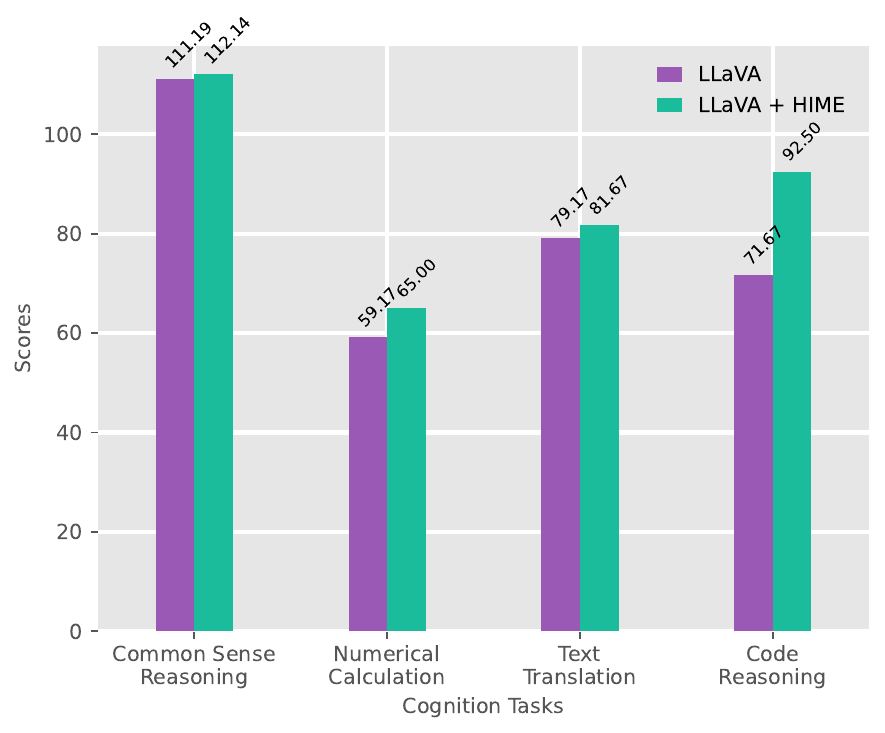}
    \vspace{-2em}
  \caption{Results from the baseline LLaVA and our approach HIME  on the cognition tasks.}
  \label{fig:gpt4v-eval}
\end{figure}

\textbf{Baselines:} To evaluate the effectiveness and generalisation of our approach in reducing the object hallucination, we utilise three representative LVLMs, including LLaVA-1.5 \cite{liu2023visual}, MiniGPT-4 \cite{zhu2023minigpt}, and mPLUG-Owl2 \cite{ye2024mplug}, and two advanced LVLMs with different LLM architecture other than LLaMA Qwen2-VL-8B-Instruct \cite{wang2024qwen2}, and Qwen3-VL-8B-Instruct \cite{Qwen3-VL}. Check the appendix for more implementation details.

\noindent\textbf{Datasets:}  
We evaluate our method on the commonly used datasets for LVLMs hallucination mitigation: CHAIR \cite{rohrbach2018object}, LLaVA-Bench \footnote{https://huggingface.co/datasets/liuhaotian/llavabench-in-the-wild}, and MME \cite{yin2024survey}, due to page limit we moved datasets details to appendix.

\section{Evaluation and Results}
This section reports experimental results demonstrating our Hallucination Insensitivity Model Editing in mitigating the object hallucination in LVLMs.

\subsection{Results on CHAIR}
As shown in Table \ref{tab:chair_results}, HIME achieves significant hallucination reduction for open-ended generation, outperforming the existing inference-time, contrastive decoding, and model editing approaches across the three models LLaVA1.5, MiiniGPT-4, and mPLUG-Owl2. Moreover, our approach, HIME, improves the state-of-the-art Qwen3-VL-8B-Instruct, and Qwen2-VL-8B-Instruct models' ability to mitigate hallucination, while preserving models' implicit knowledge, as shown in Table \ref{tab:qwen_chair_results}.
The CHAIR\(_s\) (C\(_s\)) metric is very critical as a caption is counted as an error if it contains even a single hallucinated object, regardless of how many correct objects it mentions. Consequently, a substantial improvement in CHAIR\(_s\) indicates that our approach is effective at eliminating the remaining hallucinations and making more captions entirely hallucination-free.

\begin{figure}[t]
  \centering
    \centering
    \includegraphics[width=\linewidth]{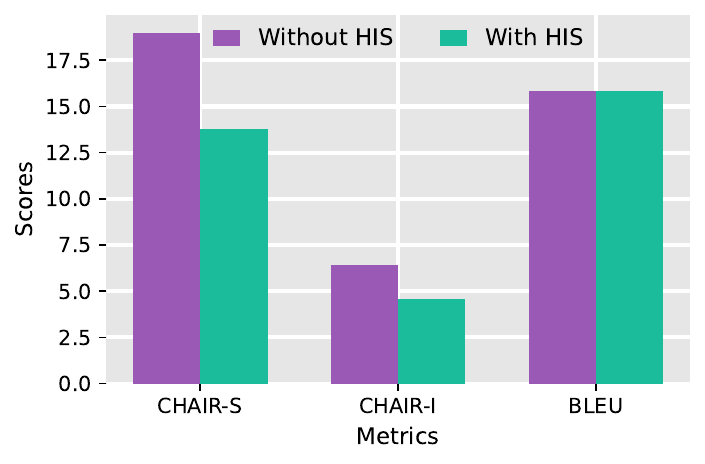}
  \vspace{-2em}
  \caption{Ablation study evaluation results for Hallucination Insensitivity Score (HIS) on LLaVA-1.5. Lower CHAIR$_s$, and CHAIR$_i$ is better, while higher BLUE is better.}
  \label{fig:kl-ablation}
\end{figure}

\subsection{Results on MME}
To evaluate the overall performance of the proposed method, we report the results of the LLaVA-1.5-7B model as a representative case. Table \ref{tab:mme-perception} shows that employing HIME consistently boosts the performance of the baseline LLaVA-1.5 model on perception tasks such as Count, Position, and Celebrity. Notably, HIME also achieves superior results outperforming the state-of-the-art approach, Nullu, with a large margin in all of the tasks, except the Colour task. HIME consistently boosts performance on perception-based tasks without degrading the models’ original recognition abilities. We attribute this to HIME’s ability to reduce statistical biases and language priors, yielding more visually grounded reasoning and, in turn, stronger overall perception. Moreover, we employed our approach with two other models MiniGPT-4 and mPLUG-Owl2, check the appendix, HIME has boosted the baseline performance significantly.

\begin{figure}[t]
  \centering
    \centering
    \includegraphics[width=\linewidth]{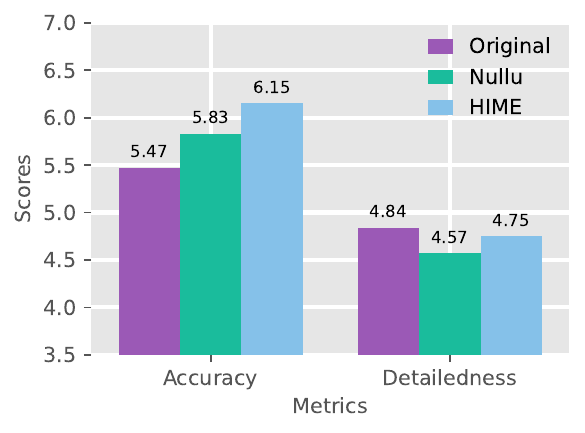}
    \vspace{-2em}
  \caption{Results from a GPT-4V–assisted evaluation of open-ended generation. Accuracy reflects how well the response matches the image, while Detailedness captures the richness of relevant detail. Both scores are reported on a 10-point scale. on LLaVA-1.5.}
  \label{fig:hime_gpt4v_eval_acc_details}
\end{figure}

\subsection{GPT-4V Aided Evaluation on LLaVA-Bench.}
Our analysis was broadened to include the open-ended captioning tasks within the LLaVA-Bench \cite{liu2024improved}, leveraging the capabilities of the recently released GPT-4V \footnote{https://openai.com/research/gpt-4v-system-card} following \cite{yang2025nullu,leng2024mitigatingvcd}. Figure \ref{fig:hime_gpt4v_eval_acc_details} shows a consistent enhancement in HIME's performance over the state-of-the-art and the baseline LLaVA model. The observed improvement in accuracy points to HIME’s capability to reduce hallucinations efficiently. Simultaneously, HIME's presents a better detailedness of the responses than Nullu but slightly less than the baseline. More qualitative studies are included in the appendix.

\noindent\textbf{Hallucination Sensitivity Score:} measures, for each layer, how sensitive attention is to factual versus hallucinated captions across the full generation process. We then use it to softly weight the magnitude of layer updates during editing. Importantly, it is not a tunable hyperparameter; it is computed once and requires no optimization. Nevertheless, we include an ablation to assess the impact of using this metric during model editing. As illustrated in Figure \ref{fig:kl-ablation}, applying smooth, layer-wise updates markedly improves performance. This suggests that rigid weight edits can discard valuable prior knowledge, which is essential for accurate generation.

\section{Conclusion}

In this work, we tackled object hallucination in LVLMs by proposing Hallucination Insensitivity Model Editing (HIME). Based on a systematic analysis of layer-wise behavior across LVLMs built on Qwen, LLaMA, and Vicuna, we showed that hallucination susceptibility varies across layers. We introduced the Hallucination Insensitivity Score (HIS) to quantify this layer-wise sensitivity and to guide targeted intervention. Building on HIS, HIME performs layer-adaptive weight editing by selectively modifying models' weights to suppress hallucinations while preserving the model’s pre-trained knowledge. Extensive evaluations demonstrate consistent hallucination reduction across benchmarks, without introducing additional parameters, inference-time latency, or computational overhead.

\bibliographystyle{named}
\bibliography{ijcai26}

\clearpage
\appendix

\twocolumn[
\begin{center}
{\LARGE\bfseries Supplementary Material for\\
HIME: Mitigating Object Hallucinations in LVLMs via Hallucination Insensitivity Model Editing\par}
\vspace{4mm}
\end{center}
]

\setcounter{equation}{0}

\vspace{5em}

\section{LVLM Hallucination Insensitivity Score}

\begin{figure*}[t!]
    \centering
\includegraphics[width=1\linewidth]{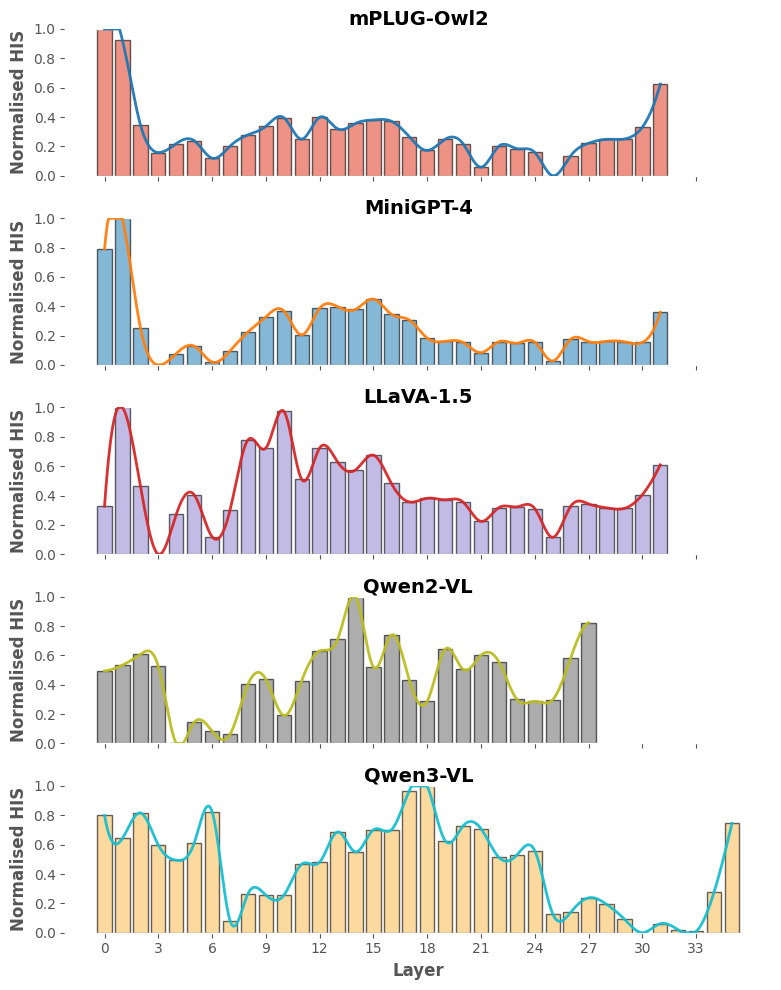}
    \vspace{-1em}
    \caption{Layer-wise distribution of the proposed normalised Hallucination Insensitivity Score (HIS) across five widely used LVLMs: mPLUG-Owl2, MiniGPT-4, LLaVA-1.5, Qwen2-vl, and Qwen3-VL. Despite architectural differences, all models exhibit similar response patterns:  layers around the middle show lower HIS values, indicating higher susceptibility to hallucinations, while deeper/early layers are comparatively more robust. Highlighting HIS as a principled indicator for identifying hallucination-prone layers and guiding targeted model editing }
    \label{fig:HIS_all_models}
\end{figure*}

As shown in Figure \ref{fig:HIS_all_models}, we extend the analysis with layer-wise distribution of the Hallucination Insensitivity Score (HIS) across three widely used LLMs backbones Vicuna, LlaMA-2, representing five widely used LVLMs: LLaVA-1.5, MiniGP), mPLUG-Owl2, Qwen2-VL, and recently proposed Qwen3-VL. Despite architectural differences, the models exhibit broadly consistent layer-wise trends, with certain layers appearing more susceptible to hallucination while others remain comparatively robust. These observations support HIS as a principled signal for identifying hallucination-prone layers and guiding targeted model editing. Note that MiniGPT-4, LLaVA-1.5, and mPLUG-Owl2 employ 32 decoder layers, whereas Qwen3-VL uses 36 layers, and Qwen2-Vl uses 28 layers.

\subsection{Ablation Study - Extended}

\textbf{Effects of editing layers:} HIME has two primary parameters: the set of layer indices to edit {\(l\)} and the number of top-\(k\) singular vectors in the null space selected. We evaluate various settings of these hyperparameters on LLaVA-1.5-7B using the CHAIR benchmark. As shown in Table \ref{tab:ablation-game}, the number of edited layers has an impact on the model's overall performance, which varies across models. Thus, in this ablation study, to assess the impact of \(l\) on the model performance, we fixed \(k\) to 5. 

\noindent As summarised in Table \ref{tab:ablation-game}, editing the later layers \(l \in \{20, 32\}\) boosts LLaVA's overall performance. This aligns with evidence that, during the late decoding phase, the model attends less to visual information and becomes prone to hallucination\cite{li2025hidden}. Our ablations also show the choice of \(k\) singular vectors is pivotal: we fix the edited layers to \(l \in \{20, 32\}\) and vary \(k\) to study its impact on the weight update.

\begin{table}[H]
\centering
\footnotesize
\setlength{\tabcolsep}{4pt}
\renewcommand{\arraystretch}{1.05}
\resizebox{\columnwidth}{!}{%
\begin{tabular}{lccc || c lccc}
\toprule
$\{\ell\}$ & \textbf{C}$_S$ & \textbf{C}$_I$ & \textbf{BLEU} & $k$ & \textbf{C}$_S$ & \textbf{C}$_I$ & \textbf{BLEU} \\
\midrule
5--32 & 18.6 & 6.44 & 15.15 & 2  & 20.2 & 6.53 & 15.58\\
10--32 &18.2  & 5.95 & 15.16 & 5  & \textbf{13.8} & \textbf{4.56} & \textbf{15.82}  \\
20--32 & \textbf{13.8} & \textbf{4.56} & \textbf{15.82} & 10  & 17.00 & 5.39 & 15.65 \\
30--32 & 15.6 & 4.99 & 15.62 & 32 & 15.8 & 5.5 & 14.29 \\
\bottomrule
\end{tabular}%
}
\caption{Ablation study of the hyper-parameters in \textsc{HIME}. On the left side, we use $k=5$ and change the layers range to be edited. On the right side, we use $l=20-32$ and change the top-ranked vectors used for weight editing.}
\label{tab:ablation-game}
\end{table}

\section{Datasets and Implementation Details}

\subsection{Datasets}

\noindent\textbf{CHAIR}: Caption Hallucination Assessment with Image Relevance measures the rate of object hallucination in the image captions by comparing generated objects with ground-truth objects. CHAIRS evaluates caption accuracy by comparing generated objects against ground-truth labels, with hallucinations defined as objects present in captions but absent from ground truth. CHAIRS assesses two levels of hallucinations: sentence-level $\text{CHAIR}_S = \dfrac{\lvert\{\text{caption w/hallucinated objects}\}\rvert}{\lvert\{\text{caption}\}\rvert}$ and instance-level $\text{CHAIR}_I=\dfrac{\lvert\{\text{hallucinated object}\}\rvert}{\lvert\{\text{object}\}\rvert}$.
Following previous settings \cite{chen2024halc,huang2024opera,yin2024woodpecker,yang2025nullu}, we randomly sample 500 images from the COCO 2014 validation set \cite{lin2014microsoft}, and use the prompt \texttt{"Please describe this image in detail."}.

\noindent\textbf{MME:} It serves as a comprehensive benchmark for evaluating LVLMs across multiple dimensions, encompassing ten perception-oriented subtasks and four cognition-focused ones.

\noindent\textbf{LLaVA-Bench: \cite{liu2024improved}} It contains 24 images paired with 60 questions covering diverse contexts—indoor and outdoor scenes, memes, paintings, and sketches. These descriptions are detailed, manually curated, and carefully designed. It probes an LVLM’s ability to handle more challenging multimodal tasks.

\subsection{Implementation Details}
This section describes how we implement the evaluated LVLMs and the strategies used to mitigate OH. The overall experimental configuration is summarized in \ref{tab:lvlm_hparams}. In contrast to standard greedy decoding—which picks the single most likely token at each step—beam search keeps a fixed set of candidate sequences (beams) and ranks them by the cumulative log-probabilities of their previous tokens \(y_{<t}\) . In our setup, we use beam search with \textit{num-beams = 3}, retaining three candidates at every step. Both baselines are implemented using the default code from the Hugging Face Transformers library \footnote{https://huggingface.co/docs/transformers}.

The complete hyperparameters for HIME across different LVLMs in our experiments are as follows. Specifically, there are three major hyper-parameters that can be actively adjusted to optimize HIME’s effectiveness across different models:

\begin{enumerate}
    \item Top-\(k\) singular vectors: The number of top-\(k\) singular vectors is crucial and it varies by model. We selected 5 for LLaVA-1.5 on CHAIR. Also, we used the top 32 singular vectors for the on mPLUG-Owl2 the two datasets. For MiniGPT-4, we employed 20 on CHAIR. In addition, for QWen2-VL-8B-Instruct, we used $k=4$ and $k=8$ for QWen3-VL-8B-Instruct.

    \item Edited Layers \(l\): denote the subset of LVLM layers whose weights are updated. Unlike Nullu \cite{yang2025nullu} approach, which restricts edits to \(l \in (16, 32)\), our method, HIME, modifies a broader span of layers and, in some models (e.g., mPLUG-Owl2) updates all layers, achieving superior results and outperforming all other existing methods. We attribute this to HIME's smooth weight adjustments, in contrast to Nullu's stricter, hard updates.

    \item Num-beam: generally speaking, in LVLMs, the number of Beams controls beam search size during each decoding step. The model keeps the top-\(k\) partial sequences (the “beams”) instead of only the single best one. In our experiments, we used different values across the models. Specifically, we set Num-beams = 1 for the mPLUG-Owl2 model, Num-beams = 2 for QWen3-VL-8B-Instruct, QWen2-VL-8B-Instruct, MiniGPT-4, and finally Num-beams = 3 for the LLaVA-1.5 model.
\end{enumerate}

\begin{table}[H]
  \centering
  \footnotesize
  \setlength{\tabcolsep}{3pt}
  \renewcommand{\arraystretch}{1.05}
  \begin{tabularx}{\columnwidth}{>{\raggedright\arraybackslash}X >{\centering\arraybackslash}c}
    \toprule
    \textbf{Parameter} & \textbf{Value} \\
    \midrule
    Do-sample & False \\
    Num-beams (beam search) & 3 \\
    Max new tokens (CHAIR) & 64 \\
    Max new tokens (MME) & 64 \\
    Max new tokens (LLaVA-Bench) & 1024 \\
    \bottomrule
  \end{tabularx}
  \caption{Hyperparameters used for LVLM decoding.}
  \label{tab:lvlm_hparams}
\end{table}

\begin{figure*}[t]
  \centering
  \begin{minipage}[t]{\columnwidth}
    \centering
    \includegraphics[width=\linewidth]{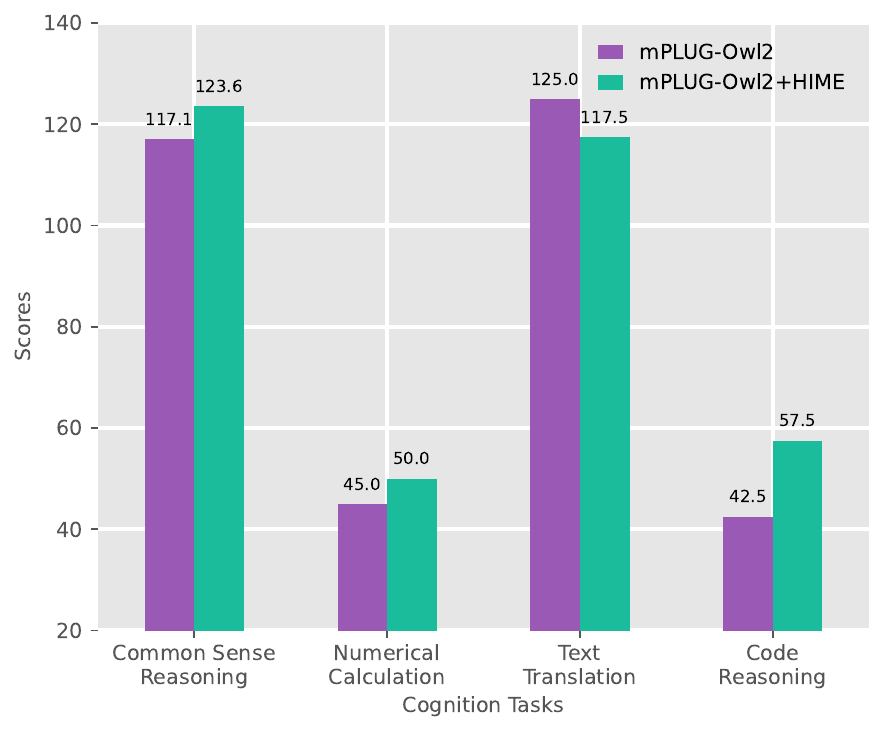}
    \vspace{-1.2em}
    \caption*{(a) mPLUG-Owl2}
  \end{minipage}\hfill
  \begin{minipage}[t]{\columnwidth}
    \centering
    \includegraphics[width=\linewidth]{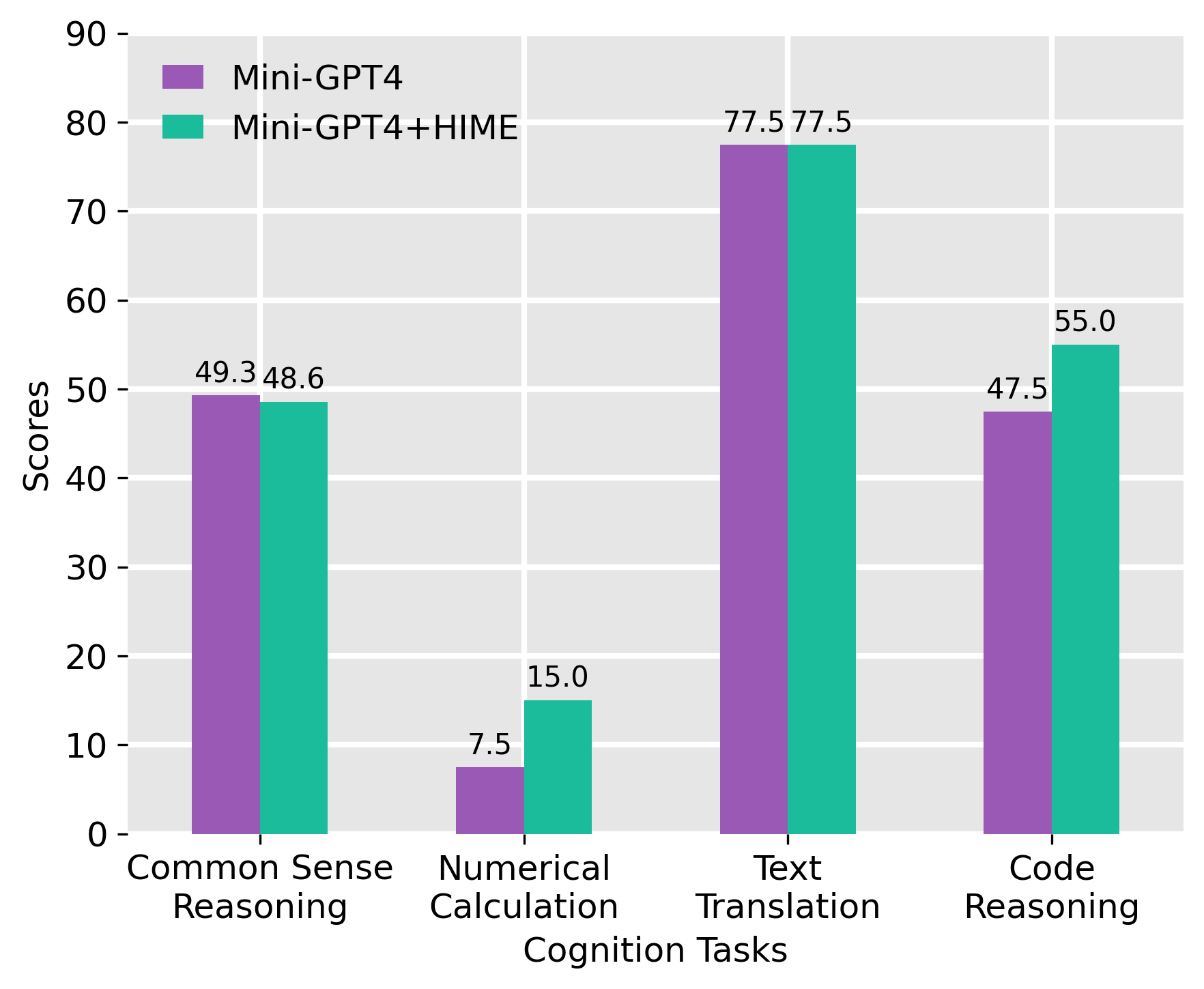}
    \vspace{-1.2em}
    \caption*{(b) MiniGPT-4}
  \end{minipage}

  \vspace{-0.8em}
  \caption{Cognition-task results for (a) mPLUG-Owl2 and (b) MiniGPT-4: baseline vs.\ HIME.}
  \label{fig:cognition_tasks_hime}
\end{figure*}

\section{Extended Evaluation}
In this section, we further evaluate \textsc{HIME} on the MME benchmark by comparing against two widely used LVLM baselines, MiniGPT-4 and mPLUG-Owl-2.

\subsection{MME}

\paragraph{MME General cognition.}
Figure~\ref{fig:cognition_tasks_hime}(b) shows that applying \textsc{HIME} to MiniGPT-4 does not impair general cognition; instead, it improves it.
While \textit{Text Translation} remains unchanged and \textit{Common Sense Reasoning} stays essentially stable, \textsc{HIME} substantially boosts more structured skills such as \textit{Numerical Calculation} and \textit{Code Reasoning}, yielding an overall increase of about $+7.9\%$ in the average cognition score.
A similar trend is observed for mPLUG-Owl2 (Fig.~\ref{fig:cognition_tasks_hime}(a)): \textsc{HIME} improves \textit{Common Sense Reasoning} and \textit{Numerical Calculation}, and delivers the largest gain on \textit{Code Reasoning}, with only a moderate drop on \textit{Text Translation}; overall, the average cognition score increases from $82.4$ to $87.2$ ($+5.8\%$), Fig.~\ref{fig:cognition_tasks_hime}.

\begin{figure*}[t]
  \centering

  \includegraphics[width=\linewidth]{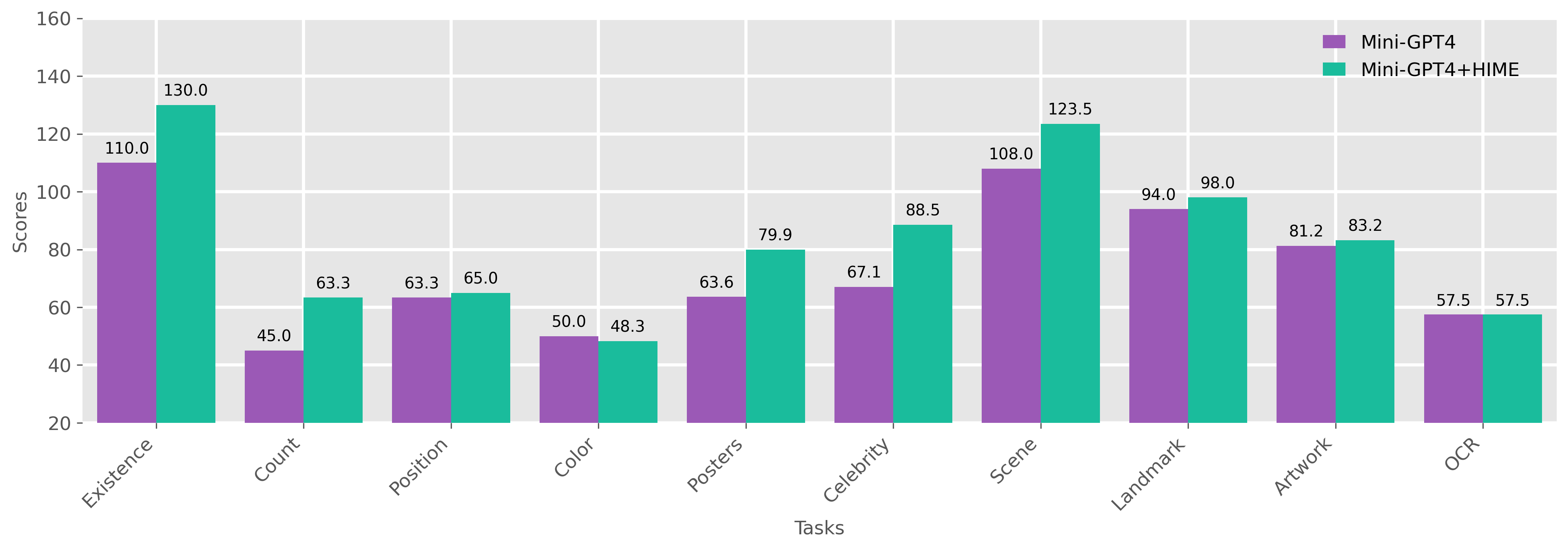}
  \vspace{-0.6em}
  \caption*{(a) MiniGPT-4}

  \vspace{0.8em} 

  \includegraphics[width=\linewidth]{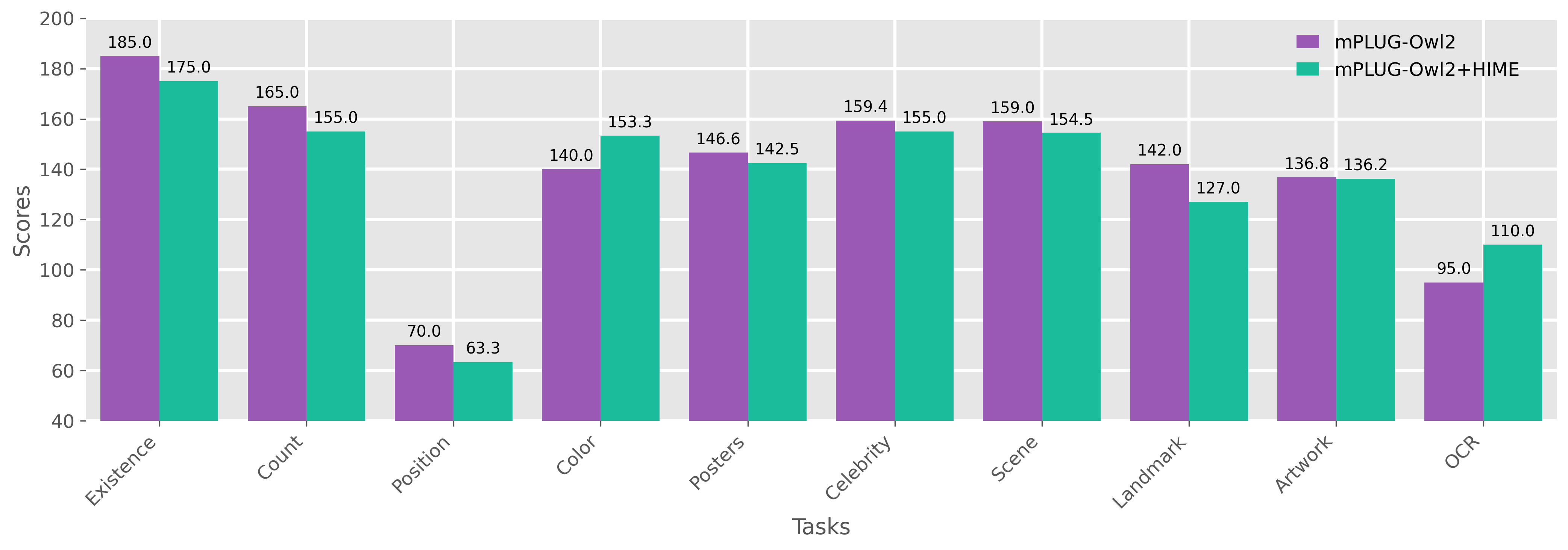}
  \vspace{-0.6em}
  \caption*{(b) mPLUG-Owl2}

  \vspace{-0.8em}
  \caption{MME perception-task results comparing the baseline and HIME for (a) MiniGPT-4 and (b) mPLUG-Owl2.}
  \label{fig:mme_perception_hime}
\end{figure*}

\paragraph{MME perception.}
Figure~\ref{fig:mme_perception_hime}(a) indicates that \textsc{HIME} consistently strengthens MiniGPT-4 on the MME perception subsets:
most categories increase noticeably (e.g., existence/counting and several recognition-oriented subsets), leading to an average gain of roughly $+13.2\%$ across perception tasks, with only a small regression in color-related judgments and no degradation in OCR.
This suggests that suppressing hallucination-prone directions can \emph{improve} visual grounding rather than merely filtering outputs.
For mPLUG-Owl2 (Fig.~\ref{fig:mme_perception_hime}(b)), \textsc{HIME} largely preserves the overall perception profile, with modest decreases in several categories but clear gains on text-centric grounding (OCR) and attribute-level recognition (e.g., color).
Taken together, these results support the central claim of \textsc{HIME}: layer-adaptive editing guided by hallucination sensitivity can mitigate hallucination-related behavior while maintaining—and in many cases enhancing—core reasoning and perceptual capabilities, without introducing inference-time overhead.

\subsection{LLaVA-Bench and GPT-4V prompt}

\begin{figure*}[t]
  \centering
    \centering
    \includegraphics[width=\linewidth]{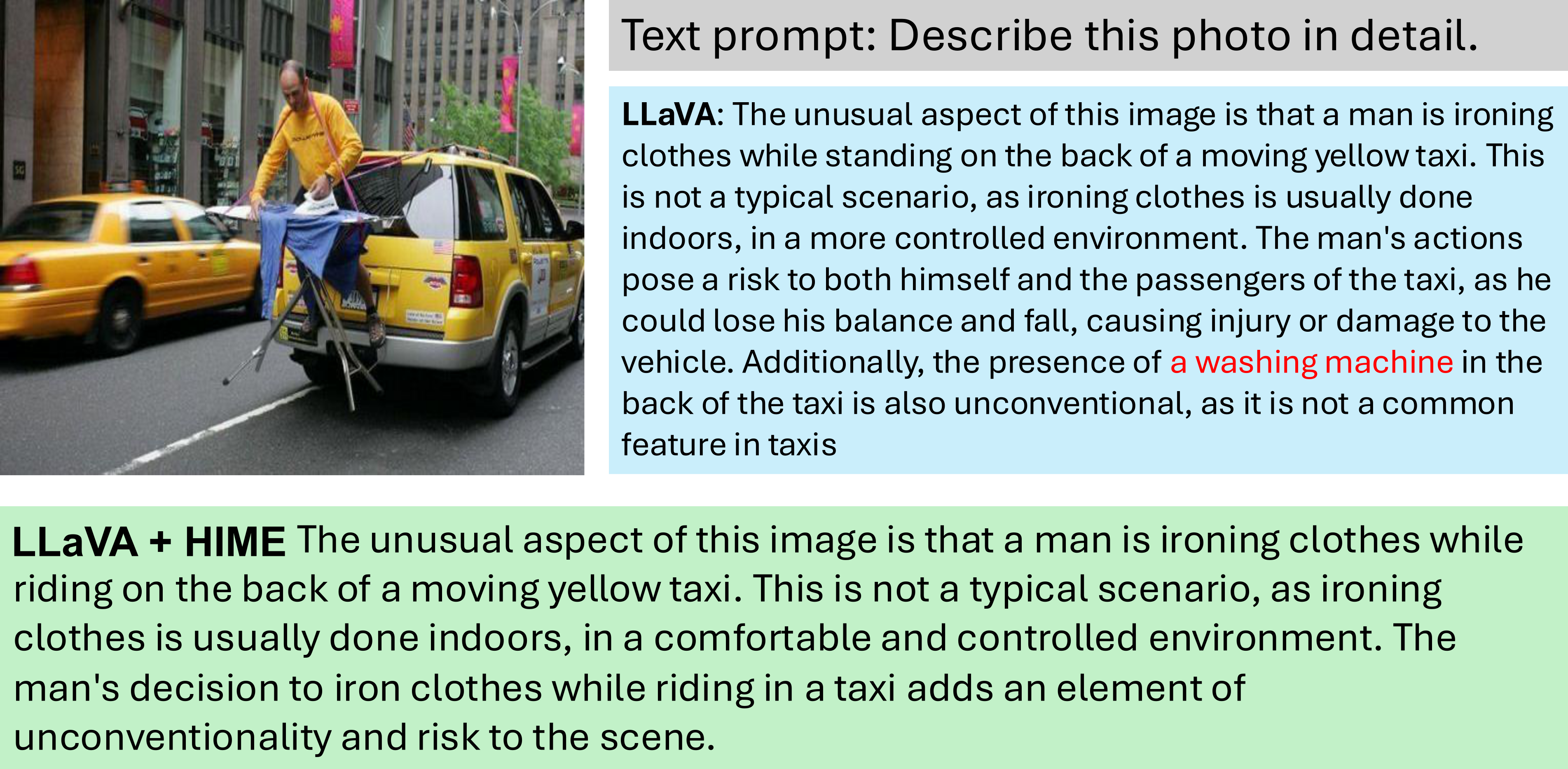}
  \caption{Illustration of hallucination correction by our proposed HIME with a sample from LLaVA-Bench using LLaVA-1.5-7B. We highlight the hallucinated objects from the original model in \textcolor{red}{red}.}
  \label{fig:llava-bench-case-study}
\end{figure*}

As we use LLaVA-Bench~\cite{liu2024improved} for qualitative comparison via GPT-4V–aided evaluation, this section describes the evaluation prompt in detail. Following~\cite{kim2023exposing}, GPT-4V scores LVLM outputs based on both factual correctness and the level of descriptive detail. The full prompt template is provided in Table \ref{tab:gpt4v_prompt}. For each test case, we first obtain responses from two LVLMs, insert each one into the template by replacing the {Response} field, and submit the resulting prompts to GPT-4V for scoring. We then parse GPT-4V’s returned scores and justifications to compare the models in terms of accuracy and detailedness. An end-to-end example of this procedure is shown in Table \ref{tab:llava_bench_gpt4v_eval_example}. Another qualitative example 
shown in Figure \ref{fig:llava-bench-case-study}, the original LLaVA model hallucinates "Washing Machine" frequently co-occurring to "ironing clothes", which is removed and mitigated with our approach.

\begin{table*}[t]
\centering

\begin{tcolorbox}[
  width=\textwidth,
  boxrule=0.9pt,
  colframe=black,
  colback=white,
  arc=3mm,
  left=5mm,right=5mm,top=4mm,bottom=4mm
]
\small
\setlength{\parindent}{0pt}

\rule{\linewidth}{0.4pt}\vspace{0.8em}

\textbf{Description:}\\
AI that scores image description accuracy and detailedness.

\vspace{1.0em}
\rule{\linewidth}{0.4pt}\vspace{0.8em}

\textbf{Instructions:}\\
You are an AI designed to evaluate and score the performance of two AI assistants in describing a given image. 
    Your primary focus is on the accuracy and detailedness of their descriptions. 
    You will assess the accuracy by checking for hallucinations - any part of the description that is inconsistent with the image content. 
    For detailedness, you will consider how rich the response is in necessary details, excluding any hallucinated parts.
    You will provide scores on a scale from 1 to 10 for each assistant separately, based on these criteria.
    After scoring, you will offer an explanation for your evaluation, ensuring it is free from bias and not influenced by the order of presentation of the responses.

\vspace{1.0em}
\textbf{Input format:}

\vspace{0.6em}
[Assistant 1]\\
\{Response 1\}\par
[End of Assistant 1]

\vspace{0.8em}
[Assistant 2]\\
\{Response 2\}\par
[End of Assistant 2]

\vspace{1.0em}
\textbf{Output format:}

\vspace{0.6em}
Accuracy:\\
Scores of the two answers:\\
Reason:

\vspace{0.8em}
Detailedness:\\
Scores of the two answers:\\
Reason:

\vspace{1.0em}
\rule{\linewidth}{0.4pt}
\end{tcolorbox}
  \caption{The prompt used to evaluate the performance of our proposed method using GPT-4V evaluator.}
  \label{tab:gpt4v_prompt}
\end{table*}

\begin{table*}[t]
\centering

\begin{tcolorbox}[evalbox]

{\color{blue}\bfseries Input:}\par
\vspace{4pt}
\includegraphics[width=0.22\textwidth]{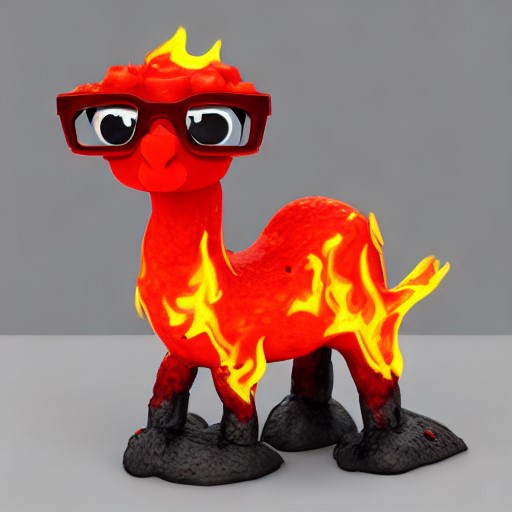}\par
\vspace{8pt}

{\color{blue}\bfseries [Assistant 1]}\par
The LLaVA logo features a red and black color scheme, with a large red fire-like element in the center. The fire element is surrounded by a black background, giving it a striking and eye-catching appearance. The design of the logo is reminiscent of a fire hydrant, which is a common symbol for fire safety and protection. The combination of the red and black colors, along with the fire-like element, creates a bold and memorable visual representation of the LLaVA brand. The design effectively communicates the brand's focus on language and vision, while also evoking a sense of warmth and safety.\par
{\color{blue}\bfseries [End of Assistant 1]}\par
\vspace{8pt}

{\color{blue}\bfseries [Assistant 2]}\par
The LLaVA logo features a unique design that combines a fire-like element with a robotic or futuristic appearance. The fire element is represented by the flames surrounding the robotic figure, which gives the logo a dynamic and eye-catching look. The robotic figure itself is depicted as a red, glasses-wearing animal, possibly a horse or a dog, with a pair of glasses. This combination of elements creates a visually striking and memorable logo that represents the innovative technology and intelligence behind LLaVA. The design effectively communicates the idea of a cutting-edge, intelligent assistant that can learn and adapt to various tasks and situations.\par
{\color{blue}\bfseries [End of Assistant 2]}\par
\vspace{10pt}

{\color{blue}\bfseries GPT4-V Output:}\par
\vspace{6pt}

{\bfseries Accuracy:}\par
{\bfseries Scores of the two answers:}\par
Assistant 1: 3\par
Assistant 2: 9\par
{\bfseries Reason:}\par
Assistant 1's description contains several inaccuracies. It mentions a fire hydrant and a focus on language and vision, which are not present in the image. The description does not accurately reflect the content of the image. \par

Assistant 2 accurately describes the image, noting the fire-like elements, the red color, the glasses, and the animal-like figure. It correctly identifies the key features without adding incorrect details.\par
\vspace{10pt}

{\bfseries Detailedness:}\par
{\bfseries Scores of the two answers:}\par
Assistant 1: 2\par
Assistant 2: 8\par
{\bfseries Reason:}\par
Assistant 1 lacks detailedness due to its inaccuracies and failure to describe the actual elements present in the image. It misses key features like the glasses and the animal-like figure.\par
Assistant 2 provides a detailed description, capturing the unique aspects of the image, such as the flames, the glasses, and the animal-like appearance. It offers a comprehensive view of the image's elements.\par

\end{tcolorbox}

\caption{Evaluation example for the image-description evaluator with GPT-4V.}
\label{tab:llava_bench_gpt4v_eval_example}
\end{table*}

\end{document}